\pdfoutput=1 
\documentclass[runningheads]{llncs}
\usepackage{graphicx}
\usepackage{scrextend}
\usepackage{amsmath,amssymb} 
\usepackage[usenames, dvipsnames]{xcolor}
 \usepackage{algorithm}
\usepackage{algpseudocode}
\usepackage{multirow}
\usepackage{tabularx}
\usepackage[width=122mm,left=12mm,paperwidth=146mm,height=193mm,top=12mm,paperheight=217mm]{geometry}
\begin{document}
\pagestyle{headings}

\mainmatter
\def\ECCV18SubNumber{2583}  

\title{DeepJDOT: Deep Joint Distribution Optimal Transport for Unsupervised Domain Adaptation}

\titlerunning{DeepJDOT}

\authorrunning{B.B. Damodaran et al.}

\author{Bharath Bhushan Damodaran$^1$\thanks{authors contributed equally}\,, Benjamin Kellenberger$^{2*}$, R\'emi Flamary$^3$, Devis Tuia$^2$, Nicolas Courty$^1$}


\institute{Universit\'e de Bretagne Sud, IRISA, UMR 6074, CNRS, France
\and
Wageningen University, the Netherlands
\and
Universit\'e C\^ote d'Azur, OCA, UMR 7293, CNRS, Laboratoire Lagrange, France\\ \texttt{\{bharath-bhushan.damodaran@irisa.fr, benjamin.kellenberger@wur.nl\}}}

\maketitle

\begin{abstract}
In computer vision, one is often confronted with problems of domain shifts, which occur when one applies a classifier trained on a source dataset to target data sharing similar characteristics (e.g. same classes), but also different latent data structures (e.g. different acquisition conditions). In such a situation, the model will perform poorly on the new data, since the classifier is specialized to recognize visual cues specific to the source domain. In this work we explore a solution, named DeepJDOT, to tackle this problem: through a measure of discrepancy on joint deep representations/labels based on optimal transport, we not only learn new data representations aligned between the source and target domain, but also simultaneously preserve the discriminative information used by the classifier.
We applied DeepJDOT to a series of visual recognition tasks, where it compares favorably against state-of-the-art deep domain adaptation methods.
\keywords{Deep Domain Adaptation, Optimal Transport}
\end{abstract}

\section{Introduction}
\label{sec:Int}
The ability to generalize across datasets is one of the holy grails of computer vision. Designing models that can perform well on datasets sharing similar characteristics such as classes, but also presenting different underlying data structures (for instance different backgrounds, colorspaces, or acquired with different devices) is key in applications where labels are scarce or expensive to obtain. However, traditional learning machines struggle in performing well out of the datasets (or \emph{domains}) they have been trained with. This is because models generally assume that both training (or \emph{source}) and test (or \emph{target}) data are issued from the same generating process. In vision problems, factors such as objects position, illumination, number of channels or seasonality break this assumption and call for adaptation strategies able to compensate for such shifts, or \emph{domain adaptation} strategies~\cite{Pat15}.

In a first rough subdivision, domain adaptation strategies can be separated into \emph{unsupervised} and \emph{semi-supervised} domain adaptation: the former assumes that no labels are available in the target domain, while the latter assumes the presence of a few labeled instances in the target domain that can be used as reference points for the adaptation. In this paper, we propose a contribution for the former, more challenging case. 
Let $\mathbf{x}^s \in \mathbb{X}^S$ be the source domain examples with the associated labels $y^s \in \mathbb{Y}^S$. Similarly, let $\mathbf{x}^t \in \mathbb{X}^T$ be the target domain images, but with unknown labels. The goal of the unsupervised domain adaptation is to learn the classifier $f$ in the target domain by leveraging the information from the source domain. To this end, we have access to a source domain dataset $\{\mathbf{x}^s_i,y^s \}_{i=1,\dots,n^s}$ and a target domain dataset $\{\mathbf{x}^t_i \}_{i=1,\dots,n^t}$ with only observations and no labels.

Early unsupervised domain adaptation research tackled the problem as the one of finding a common representation between the domains, or a latent space, where a single classifier can be used independently from the datapoint's origin~\cite{saenko10,gopalan11}. In~\cite{Cou17}, the authors propose to use discrete optimal transport to match the shifted marginal distributions of the two domains under constraints of class regularity in the source. In~\cite{courty2017joint} a similar logic is used, but the joint distributions are aligned directly using a coupling accounting for the marginals and the class-conditional distributions shift \emph{jointly}. However, the method has two drawbacks, for which we propose solutions in this paper: 1) first, the JDOT method in \cite{courty2017joint} scales poorly, as it must solve a $n_1 \times n_2$ coupling, where $n_1$ and $n_2$ are the samples to be aligned; 2) secondly, the optimal transport coupling $\gamma$ is computed between the input spaces (and using a $\ell_2$ distance), which is a poor representation to be aligned, since we are interested in matching more semantic representations supposed to ease the work of the classifier using them to take decisions.

We solve the two problems above by a strategy based on deep learning. On the one hand, using deep learning algorithms for domain adaptation has found an increasing interest and has shown impressive results in recent computer vision literature~\cite{deepcoral,LEL,Ganin2016,CoGAN}. On the other hand (and more importantly), a Convolutional Neural Network (CNN) offers the characteristics needed to solve our two problems:  1) by gradually adapting the optimal transport coupling along the CNN training, we obtain a scalable solution, an approximated and stochastic version of JDOT; 2) by learning the coupling in a deep layer of the CNN, we align the representation the classifier is using to take its decision, which is a more semantic representation of the classes. In summary, we learn jointly the embedding between the two domains and the classifier in a single CNN framework. We use a domain adaptation-tailored loss function based on optimal transport and therefore call our proposition \emph{Deep Joint Distribution Optimal Transportation (DeepJDOT)}.

We test DeepJDOT on a series of visual domain adaptation tasks and compare favorably against several recent state of the art competitors.

\section{Related works}
\label{sec:rw}

\paragraph{Unsupervised domain adaptation.} Unsupervised domain adaptation studies the situation where the source domain carries labeled instances, while the target domain is unlabeled, yet accessible during training~\cite{BenD07}. 
Earlier approaches consider projections aligning data spaces to each other~\cite{saenko10,Jhuo12,Hoffman_ICLR2013}, thus trying to exploit shift-invariant information to match the domains in their original (input) space. Later works extended such logic to deep learning, typically by weight sharing~\cite{deepcoral}/reconstruction~\cite{Alj16}, by adding Maximum Mean Discrepancy (MMD) and association-based losses between source and target layers~\cite{LongMMD,Lon16,AssocDA}. Other major developments focus on the inclusion of adversarial loss functions pushing the CNN to be unable to discriminate whether a sample comes from the source or the target domain~\cite{LEL,Ganin2016,Tzeng15}. Finally, the most recent works extend this adversarial logic to the use of GANs~\cite{UNIT,GenToAdapt}, for example using two GAN modules with shared weights~\cite{CoGAN}, forcing image to image architecture{s} to have similar activation distributions~\cite{I2IAdapt} or simply fooling a GAN's discriminator discerning between domains~\cite{Adda}. These adversarial image generation based methods \cite{UNIT,GenToAdapt,I2IAdapt} use a class-conditioning or cycle consistency term to learn the discriminative embedding, such that semantically similar images in both domains are projected closeby in the embedding space. Our proposed DeepJDOT uses the concept of a shared embedding for both domains~\cite{Tzeng15} and is built on a similar logic as the MMD-based methods, yet adding a clear discriminative component to the alignment: 
the proposed DeepJDOT associates representation and discriminative learning, since the optimal transport coupling ensures that distributions are matched, while {\em i)} the JDOT class loss performs source label propagation to the target samples\, and {\em ii)} the fact of learning the coupling in deep layers of the CNN ensures discrimination power.

\paragraph{Optimal transport in domain adaptation.} Optimal transport~\cite{monge1781memoire,Kantorovich42,Villani09} has been used in domain adaptation to learn the transformation between domains~\cite{Cou17,Cou14,Per16}, with associated theoretical guarantees~\cite{redko2017}. In those works, the coupling $\gamma$ is used to transport (i.e. transform) the source data samples through an estimated mapping called barycentric mapping. Then, a new classifier is trained on the transported source data representation. But those different methods can only address problems of small to medium sizes because they rely on the exact solution of the OT problem on all samples. 
Very recently, Shen {\em et al.}~\cite{Shen2018} used the Wasserstein distance as a loss in a deep learning setting to promote similarities between embedded representations using the dual formulation of the problem exposed in~\cite{arjovsky17}. However, none of those approaches considers an adaptation w.r.t. the discriminative content of the representation, as we propose in this paper.

\section{Optimal transport for domain adaptation}
\label{sec:ot}

Our proposal is based on optimal transport. After recalling the associated basic notions and its relation with domain adaptation, we detail the JDOT method~\cite{courty2017joint}, which is the starting point of our proposition.
 
\subsection{Optimal Transport}
Optimal transport~\cite{Villani09} (OT) is a theory that allows to compare probability distributions in a geometrically sound manner. It permits to work on empirical distributions and to exploit the geometry of the data embedding space. Formally, OT searches a probabilistic coupling $\gamma \in \Pi(\mu_1,\mu_2)$ between two distributions $\mu_1$ and $\mu_2$ which yields a minimal displacement cost 
  \begin{equation}
 OT_c(\mu_1,\mu_2) = \inf_{\gamma\in \Pi(\mu_1,\mu_2)} \int_{\mathcal{R}^2} c(\mathbf{x}_1,\mathbf{x}_2)d \gamma(\mathbf{x}_1,\mathbf{x}_2)
 \end{equation}
w.r.t. a given cost function $c(\mathbf{x}_1,\mathbf{x}_2)$ measuring the dissimilarity between samples $\mathbf{x}_1$ and $\mathbf{x}_2$. Here, $\Pi(\mu_1,\mu_2)$ describes the
space of joint probability distribution{s} with marginals $\mu_1$ and $\mu_2$. In a discrete setting (both distributions are empirical) this becomes:
 \begin{equation}
 OT_c(\mu_1,\mu_2) = \min_{\gamma \in \Pi(\mu_1,\mu_2)}  <\gamma, \mathbf{C}>_F,
 \label{eq:kanto}
 \end{equation}
 where $\langle\cdot, \cdot\rangle_F$ is the Frobenius dot product, $\mathbf{C} \geq 0$ is a cost matrix $\in \mathbb{R}^{n_1\times n_2}$ representing the pairwise costs $c(\mathbf{x}_i,\mathbf{x}_j)$, and $\gamma$ is a matrix of size $n_1\times n_2$ with prescribed marginals. The minimum of this optimization problem can be used as a distance between distributions, and, whenever the cost $c$ is a norm, it is referred to as the Wasserstein distance. Solving equation~\eqref{eq:kanto} is a simple linear programming problem with equality constraints, but scales super-quadratically with the size of the sample. Efficient computational schemes were proposed with entropic regularization~\cite{Cuturi13} and/or stochastic versions using the dual formulation of the problem~\cite{genevay2016,arjovsky17,seguy2018}, allowing to tackle small to middle sized problems. 

\subsection{Joint Distribution Optimal Transport}
Courty et al.~\cite{courty2017joint} proposed the joint distribution optimal transport (JDOT) method to prevent the two-steps adaptation (i.e. first adapt the representation and then learn the classifier on the adapted features) by directly learning a classifier embedded in the cost function $c$. The underlying idea is to align the joint features/labels distribution instead of only considering the features distribution. Consequently, $\mu_s$ and $\mu_t$ are measures of the product space $\mathcal{X}\times\mathcal{Y}$. The generalized cost associated to this space is expressed as a weighted combination of costs in the feature and label spaces, reading
 \begin{equation}
d \left( \mathbf{x}_i^s, \mathbf{y}_i^s;\mathbf{x}_j^t, \mathbf{y}_j^t \right) = \alpha c(\mathbf{x}_i^s,\mathbf{x}_j^t) + \lambda_t L(\mathbf{y}_i^s,\mathbf{y}_j^t)
 \label{eq:cost}
 \end{equation}
 for the $i$-th source and $j$-th target element, and where $c(\cdot,\cdot)$ is chosen as a $\ell^2_2$ distance and $L(\cdot,\cdot)$ is a classification loss (e.g. hinge or cross-entropy). Parameters $\alpha$ and $ \lambda_t$ are two scalar values weighing the contributions of distance terms. Since target labels $\mathbf{y}_j^t$ are unknown, they are replaced by a surrogate version $f(\mathbf{x}_j^t)$, which depends on a classifier $f:\mathcal{X}\rightarrow\mathcal{Y}$. Accounting for the classification loss leads to the following minimization problem:
 \begin{equation}
 \min_{f,\gamma \in \Pi(\mu_s,\mu_t)}  <\gamma, \mathbf{D}_f>_F,
 \end{equation}
 where  $\mathbf{D}_f$ depends on $f$ and gathers all the pairwise costs $d(\cdot,\cdot)$. As a by-product of this optimization problem, samples that share a common representation and a common label (through classification) are matched, yielding better discrimination. Interestingly, it is proven in~\cite{courty2017joint} that minimizing this quantity is equivalent to minimizing a learning bound on the domain adaptation problem. However, JDOT has two major drawbacks: {\em i)} on large datasets, solving for $\gamma$ becomes intractable because $\gamma$ scales quadratically in size to the number of samples; {\em ii)} the cost $c(\mathbf{x}_i^s,\mathbf{x}_j^t)$ is taken in the input space as the squared Euclidean norm on images and can be uninformative of the dissimilarity between two samples. Our proposed DeepJDOT solves those two issues by introducing a stochastic version computing only small couplings along the iterations of a CNN, and by the fact that the optimal transport is learned between the semantic representations in the deeper layers of the CNN, rather than in the image space.

\section{Proposed method}

\subsection{Deep Joint Distribution Optimal Transport(DeepJDOT)}

\begin{figure}[!t]
\centerline{
  \includegraphics[width=0.8\linewidth]{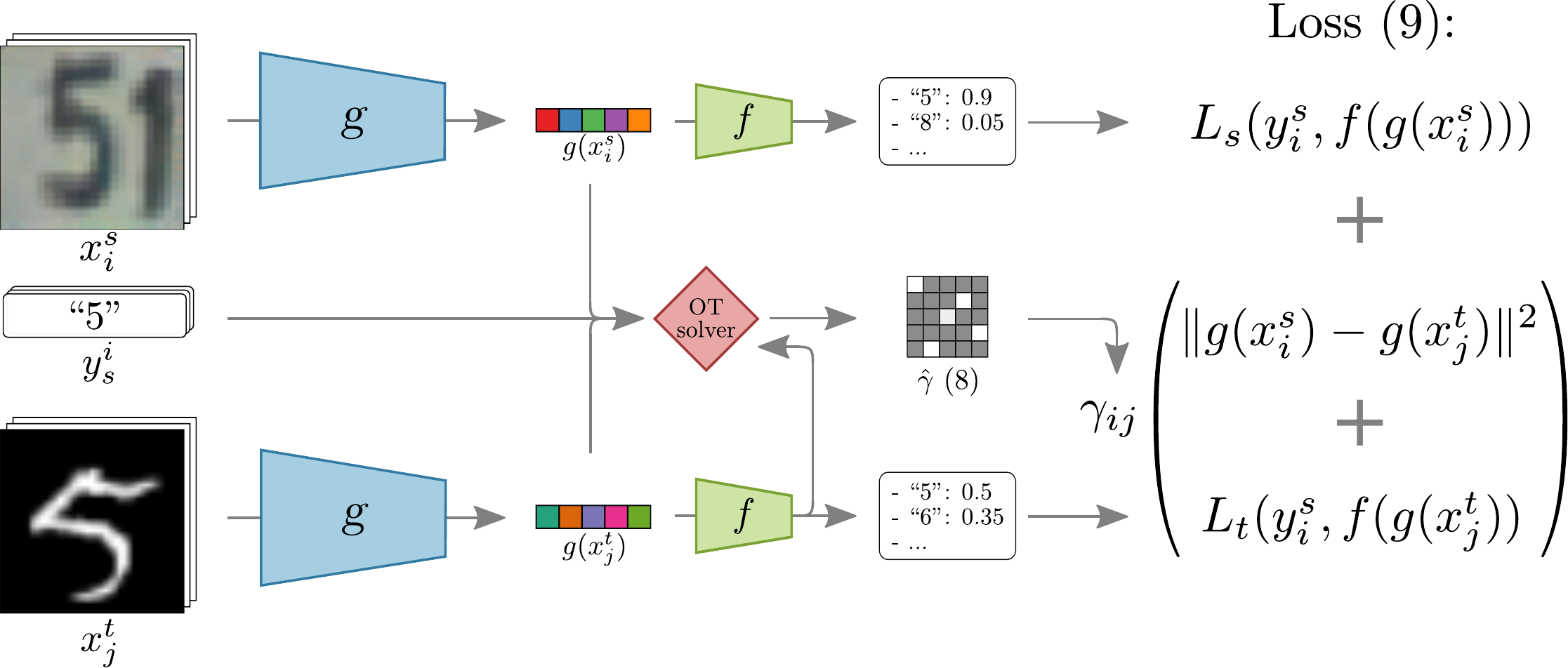}
}
  \caption{Overview of the proposed DeepJDOT method. While the structure of the feature extractor $g$ and the classifier $f$ are shared by both domains, they are represented twice to distinguish between the two domains. Both the latent representations and labels are used to compute per batch a coupling matrix $\gamma$ that is used in the global loss function.}
  \label{fig:flow}
\end{figure}

The DeepJDOT model, illustrated in Fig.~\ref{fig:flow}, is composed of two parts: an embedding function $g:\mathbf{x} \rightarrow \mathbf{z}$, where the input is mapped into the latent space $Z$, and the classifier $f: \mathbf{z} \rightarrow \mathbf{y}$, which maps the latent space to the label space on the target domain. 
The latent space can be any feature layer provided by a model, as in our case the penultimate fully connected layer of a CNN.
DeepJDOT optimizes jointly this feature space and the classifier to provide a method that performs well on the target domain. 
The solution to this problem can be achieved by minimizing the following objective function:
\begin{equation}
\min_{\gamma\in \Pi(\mu_s,\mu_t), f,g}\quad \sum_i\sum_j \gamma_{ij} d \left( g(\mathbf{x}_i^s), \mathbf{y}_i^s;g(\mathbf{x}_j^t), f(g(\mathbf{x}_j^t)) \right),
\label{eq:jdot} 
\end{equation}
where $d \left( g(\mathbf{x}_i^s), \mathbf{y}_i^s;g(\mathbf{x}_j^t), f(g(\mathbf{x}_j^t) \right) =\alpha \Vert g(x_i^s) - g(x_j^t) \Vert^2 + \lambda_t L\left(y_i^s,f(g(x_j^t))\right)$, and $\alpha$ and $\lambda_t$ are the parameters controlling the tradeoff between the two terms, as in equation~\eqref{eq:cost}. Similarly to JDOT, the first term in the loss compares the compatibility of the embeddings for the source and target domain, while the second term considers the classifier $f$ learned in the target domain and its regularity with respect to the labels available in the source. Despite similarities with the formulation of JDOT~\cite{courty2017joint}, our proposition comes with the notable difference that, in DeepJDOT, the Wasserstein distance is minimized between the joint (embedded space/label) distributions within the CNN, rather than between the original input spaces. As the deeper layers of a CNN encode both spatial and semantic information, we believe them to be more apt to describe the image content for both domains, rather than the original features that are affected by a number of factors such as illumination, pose or relative position of objects.

One can note that the formulation reported in equation~\eqref{eq:jdot} only depends on the classifier learned in the target domain. By doing so, one puts the emphasis on learning a good classifier for the target domain, and disregards the performance of the classifier when considering source samples. In recent literature, such a degradation in the source domain has been named as `\emph{catastrophic forgetting}'~\cite{Shm17,Li18}. To avoid such forgetting, one can easily re-incorporate the loss on the source domain in \eqref{eq:jdot}, leading to the final DeepJDOT objective:

 \begin{equation}
  \min_{\gamma, f,g} \frac{1}{n^s}\sum_i L_s\left(y_i^s, f(g(x_i^s))\right) + \sum_{i,j} \gamma_{ij} \left(\alpha \Vert g(x_i^s) - g(x_j^t) \Vert^2 + \lambda_t L_t\left(y_i^s,f(g(x_j^t))\right) \right).
  \label{eq:jdot+source2}
  \end{equation}
This last formulation is the optimization problem solved by DeepJDOT. However, for large sample sizes the constraint of computing a full $\gamma$ yields a computationally infeasible problem, both in terms of memory and time complexity. In the next section, we propose an approximation method based on stochastic optimization.

\subsection{Solving the optimization problem with stochastic gradients}
\label{sec:}
In this section, we describe the {approximate} optimization procedure for solving DeepJDOT. {Equation~\eqref{eq:jdot+source2} involves} two groups of variables to be optimized: the OT matrix $\gamma$ and the models $f$ and $g$. This suggest the use of an alternative minimization approach (as proposed in the original JDOT). 
Indeed,  when $\hat{g}$ and $\hat{f}$ are fixed, solving equation \eqref{eq:jdot+source2} boils down to {a standard} OT problem with associated cost matrix $C_{ij} = \alpha \Vert \hat{g}(x_i^s) - \hat{g}(x_j^t) \Vert^2 + \lambda_t  L_t\left(y_i^s,\hat{f}(\hat{g}(x_j^t))\right)$. When fixing $\hat{\gamma}$, optimizing $g$ and $f$ is a classical deep learning problem. 
However, computing the optimal coupling with the classical OT solvers is not scalable to large-scale datasets. Despite some recent development for large scale OT with general ground loss~\cite{genevay2016,seguy2018}, the model does not scale sufficiently to meet requirements of recent computer vision tasks.

Therefore, in this work we propose to solve the problem with a stochastic approximation using minibatches from both the source and target domains \cite{genevay2017sinkhorn}. This approach has two major advantages: it is scalable to large datasets and can be easily integrated in modern deep learning frameworks.
More specifically, the objective function~\eqref{eq:jdot+source2} is approximated by sampling a mini-batch of size $m$, {leading to the following optimization problem:}
 \begin{equation}
 \small
  \min_{f,g} \mathbb{E}\left[\frac{1}{m}\sum_{i=1}^{m} L_s\left(y_i^s, f(g(x_i^s)\right) +\min_{\gamma\in\Delta}\sum_{i,j}^{m} \gamma_{ij} \left(\alpha \Vert g(x_i^s) - g(x_j^t) \Vert^2 + \lambda_t L_t\left(y_i^s,f(g(x_j^t))\right) \right)\right]
  \label{eq:jdot+source_batch}
  \end{equation}

where $\mathbb{E}$ is the expected value with respect to the randomly sampled minibatches drawn from both source and target domains. The classification loss functions for the source ($L_s$) and target ($L_t$) domains can be any general class of loss functions that are twice differentiable. We opted for a traditional cross-entropy loss in both cases. Note that, as discussed in \cite{genevay2017sinkhorn}, the expected value over the minibtaches does not converge to the true OT coupling between every pair of samples, which might then lead to the appearance of connections between samples that would not have been connected in the full coupling. However, this can also be seen as a regularization that will promote sharing of the mass between neighboring samples. Finally note that we did not use the regularized version of OT as in \cite{genevay2017sinkhorn}, since it introduces an additional regularization parameter that should be cross-validated, which can make the model calibration even more complex. Still, the extension of DeepJDOT to regularized OT is straightforward and could be beneficial for high-dimensional embeddings $g$.

Consequently, we propose to obtain the stochastic update for Eq.\eqref{eq:jdot+source_batch}
as follows (and summarized in Algorithm \ref{al:deepjdot+emd}):
\begin{enumerate}
    \item With fixed CNN parameters $(\hat{g},\hat{f})$ and for every randomly drawn minibatch (of $m$ samples), obtain the coupling
    \begin{equation}
     \min_{\gamma\in \Pi(\mu_s,\mu_t)}\quad\sum_{i,j=1}^{m} \gamma_{ij} \left(\alpha \Vert \hat{g}(x_i^s) - \hat{g}(x_j^t) \Vert^2 + \lambda_t  L_t\left(y_i^s,\hat{f}(g(x_j^t))\right) \right) 
     \label{eq:jdot_gamma}
     \end{equation}
    using the network simplex flow algorithm. 
    \item With fixed coupling $\hat{\gamma}$ obtained at the previous step, update the embedding function ($g$) and classifier ($f$) with stochastic gradient update for the following loss on the minibatch:
    \begin{equation}
      \frac{1}{m}\sum_{i=1}^{m} L_s\left(y_i^s, f(g(x_i^s))\right) + \sum_{i,j=1}^{m} \hat{\gamma}_{ij} \left(\alpha \Vert g(x_i^s) - g(x_j^t) \Vert^2 + \lambda_t  L_t\left(y_i^s,f(g(x_j^t))\right) \right).
    \label{eq:jdot+fandg}
    \end{equation}
The domain alignment term aligns only the source and target samples with similar activation/labels and the sparse matrix $\hat{\gamma}$ will automatically perform label propagation between source and target samples. The classifier $f$ is simultaneously learnt in both source and target domain.
\end{enumerate}

\begin{algorithm}[!t]
  \begin{algorithmic}[1]
  \Require{
    $\mathbf{x}^s$: source domain samples, $\mathbf{x}^t$: target domain samples, $\mathbf{y}^s$: source domain labels}
  \For{each batch of source ($\mathbf{x_b}^s, \mathbf{y_b}^s$)  and target samples ($\mathbf{x_b}^t$)}
   \State fix $\hat{g}$ and $\hat{f}$, solve for $\gamma$ as in {equation} \eqref{eq:jdot_gamma}
   \State fix $\hat{\gamma}$, and update for $g$ and $f$ according to {equation} \eqref{eq:jdot+fandg}   	
    \EndFor
    \label{al:deepjdot+emd}
  \end{algorithmic}
  \caption{DeepJDOT stochastic optimization}
\end{algorithm}

\section{Experiments and Results}

We evaluate DeepJDOT on three adaptation tasks: digits classification (Section~\ref{sec:data}), the OfficeHome dataset (Section~\ref{sec:officehome}), and the Visual Domain Adaptation challenge (visDA; Section~\ref{sec:visda}). For each dataset, we first present the data, then detail the implementation and finally present and discuss the results.

\subsection{Digit classification}\label{sec:data}

\subsubsection*{Datasets} 
\begin{figure}
\centering
\includegraphics[width=\linewidth]{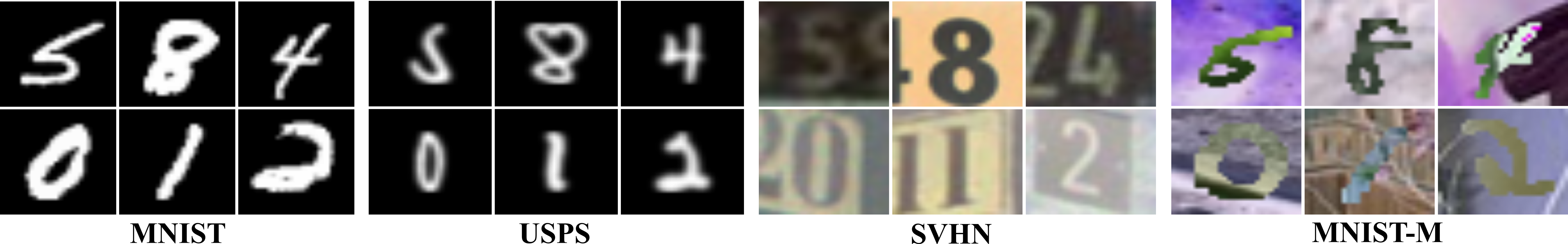}
\caption{Examples from the MNIST, USPS, SVHN and MNIST-M datasets.}
\label{fig:datasets}
\end{figure}

We consider four data sources (domains) from the digits classification field: MNIST \cite{Lecun98}, USPS \cite{Hull94}, MNIST-M, and the Street View House Numbers (SVHN) \cite{Netzer11} dataset. Each dataset involves a 10-class classification problem (retrieving numbers 0-9):

\begin{itemize}
\item[-] \emph{USPS}. The USPS datasets contains $7`291$ training and $2`007$ test grayscale images of handwritten images, each one of size $16 \times16$ pixels.
\item[-] \emph{MNIST}. The MNIST dataset contains $60`000$ training and $10`000$ testing grayscale images of size 28 $\times$ 28.
\item[-] \emph{MNIST M}. We generated the MNIST-M images by following the protocol in \cite{Ganin2016}. MNIST-M is a variation on MNIST, where the (black) background is replaced by random patches extracted from the Berkeley Segmentation Data Set (BSDS500) \cite{Arbelaez11}. The number of training and testing samples are the same as the MNIST dataset discussed above.
\item[-] \emph{SVHN}. The SVHN dataset contains house numbers extracted from Google Street View images. We used the \textit{Format2} version of SVHN, where the images are cropped into 32 $\times$ 32 pixels. Multiple digits can appear in a single image, the objective is to detect the digit in the center of the image. This dataset contains $73`212$ training images, and $26`032$ testing images of size 32$\times$ 32$\times$3. The respective examples of the each dataset is shown in Figure \ref{fig:datasets}.
\end{itemize}

\noindent The three following experiments were run (the arrow direction corresponds to the sense of the domain adaptation):
\begin{itemize}
\item[-] \emph{USPS$\leftrightarrow$MNIST}. The USPS images are zero-padded to reach the same size as MNIST dataset.
The adaptation is considered in both directions: USPS $\rightarrow$ MNIST, and MNIST $\rightarrow$ USPS.

\item[-] \emph{SVHN$\rightarrow$MNIST}. The single-channel MNIST images are replicated three times to form a gray 3 channels image, and resized to match the resolution of the SVHN images. Here, the adaptation is considered in only one direction: SVHN$\rightarrow$MNIST. Adapting SVHN images to MNIST is challenging due to the variations in the SVHN images \cite{Ganin2016}  
\item[-] \emph{MNIST$\rightarrow$MNIST-M}. MNIST is considered as the source domain and MNIST-M as the target domain. The color MNIST-M images can be easily identified by a human, however it is challenging for the CNN trained on MNIST, which is only grayscale. Again, the gray scale MNIST images are replicated three times to match the color resolution of the MNIST-M images.
\end{itemize}

\subsubsection*{Model} For all digits adaptation experiments, our embedding function $g$ is trained from scratch with six $3\times3$ convolutional layers containing 32, 32, 64, 64, 128 and 128 filters, and one fully-connected layer of 128 hidden units followed by a sigmoid nonlinearity respectively. Classifier $f$ then consists of a fully-connected layer, followed by a softmax to provide the class scores.
The Adam optimizer ($lr = 2e{-4}$) is used to update our model using mini-batch sizes of $m_S = m_T = 500$ for the two domains ($50$ samples per class in the source mini-batch). The hyper-parameters of DeepJDOT, $\alpha = 0.001$ and $\lambda_t = 0.0001$, are fixed experimentally. 

We compare DeepJDOT with the following methods: 
\begin{itemize}
\item non-adversarial discrepancy methods: DeepCORAL \cite{deepcoral}, MMD\cite{LongMMD}, DRCN\cite{DRCN}, DSN \cite{DSN}, AssocDA\cite{AssocDA}, Self-ensemble\cite{french2018}{\footnote{we report a comparison against \cite{french2018} by using minimal data augmentation (corresponding to MT+CT$^*$ in
Table 1 of \cite{french2018}). We do not compare against their full model, as they use a much heavier data augmentation and different networks.\label{ft:self-ensemble}}}, 
\item adversarial discrepancy methods: DANN\cite{Ganin2016}, ADDA\cite{Adda}, 
\item adversarial image generation methods: CoGAN\cite{CoGAN}, UNIT\cite{UNIT},  GenToAdapt\cite{GenToAdapt} and I2I Adapt\cite{I2IAdapt}. 
\end{itemize}
To ensure fair comparison, we re-implemented the most relevant competitors (CORAL, MMD, DANN, and ADDA). For the other methods, the results are directly reported from the respective articles.

\subsubsection*{Results} The performance of DeepJDOT on the four digits adaptation tasks is reported in Table~\ref{tab:results_largeCNN}. The first row (\texttt{source only}) shows the accuracies on target test data achieved with classifiers trained on source data without adaptation, and the row (\texttt{target only}) reports accuracies on the target test data achieved with classifiers trained on the target training data. This  method is considered as an upper bound for our proposed method and can be seen as our gold standard. \texttt{StochJDOT} (stochastic adaptation of JDOT) refers to the accuracy of our proposed method, when the discrepancy between source and target domain is computed with an $\ell_2$ distance in the original image space. Lastly, \texttt{DeepJDOT-source} indicates the source data accuracy, after adapting to the target domain, and can be considered a measure of catastrophic forgetting.

The experimental results show that DeepJDOT achieves accuracies comparable or higher to the current state-of-the-art methods. When the methods in the first block of Table~\ref{tab:results_largeCNN} are considered, \texttt{DeepJDOT} outperforms the competitors by large margins, with the exception of \texttt{DANN} that have similar performance on the MNIST$\rightarrow$USPS task. In the more challenging adaptation settings (SVHN$\rightarrow$MNIST and MNIST$\rightarrow$MNIST-M),
the state-of-the-art methods{\footnote{For \texttt{ADDA}\cite{Adda} in the SVHN$\rightarrow$MNIST adaptation task the accuracy is reported from the paper, as we were not able to further improve the source only accuracy \label{ft:adda}}}
 were not able to adapt well to the target domain. Next, when the methods in the second block of Table~\ref{tab:results_largeCNN} is considered, our method showed impressive performance, despite \texttt{DeepJDOT} not using any complex procedure for generating target images to perform the adaptation. 

\begin{table}[t]
\caption{Classification accuracy on the target test datasets for the digit classification tasks. \textit{Source only} and \textit{target only} refer to training on the respective datasets without domain adaptation and evaluating on the target test dataset. The accuracies reported in the first block are our own implementations, while the second block reports performances from the respective articles. \textbf{Bold} and \textit{italic} indicates the best and second best results. The last line reports the performance of DeepJDOT on the source domain.}
\label{tab:results_largeCNN} 
\resizebox{\textwidth}{!}{\begin{tabular}{c|c|c|c|c}
\hline
\multirow{2}{*}{Method}& \multicolumn{4}{c}{Adaptation:source$\rightarrow$target} \\
  & MNIST $\rightarrow$ USPS & USPS $\rightarrow$ MNIST & SVHN $\rightarrow$ MNIST  & MNIST $\rightarrow$ MNIST-M\\
\hline
Source only & 94.8  &  59.6 &60.7 &60.8\\\hline\hline
DeepCORAL \cite{deepcoral} &89.33 &91.5&59.6& 66.5\\
MMD  \cite{LongMMD}&88.5 & 73.5 &64.8&72.5 \\
DANN \cite{Ganin2016}& \textit{95.7}&  90.0&70.8&75.4 \\
ADDA \cite{Adda} & 92.4& 93.8& 76.0\footref{ft:adda} & 78.8\\
\hline
AssocDA \cite{AssocDA} &- & -& \textit{95.7} & \textit{89.5}\\
Self-ensemble\footref{ft:self-ensemble}\cite{french2018} & 88.14 & 92.35 & 93.33 & - \\
DRCN \cite{DRCN} & 91.8& 73.6&81.9&- \\
DSN \cite{DSN} &91.3& -&82.7&83.2\\

CoGAN \cite{CoGAN} & 91.2& 89.1& -&-\\
UNIT \cite{UNIT} & \textbf{95.9}& \textit{93.5}&90.5&- \\
GenToAdapt \cite{GenToAdapt} &95.3&90.8&92.4&- \\
I2I Adapt \cite{I2IAdapt} & 92.1&87.2 &80.3 &-\\

\hline
\hline
StochJDOT & 93.6&90.5 & 67.6&66.7\\
DeepJDOT (ours) & \textit{95.7}& \textbf{96.4} & \textbf{96.7} & \textbf{92.4}\\
\hline\hline
target only &95.8& 98.7& 98.7& 96.8 \\\hline
DeepJDOT-source & 98.5&94.9 & 75.7&97.8\\\hline
\end{tabular}
}
\end{table}

\subsubsection{$t$-SNE embeddings} 
\begin{figure}[tbp]
\centering
\def\wfig{0.35}
\begin{tabular}{cc@{\hskip 1cm}c}
  &\bf Source (red) VS target (blue) & \bf Class discrimination \\
  \raisebox{2.cm}{\rotatebox[origin=c]{90}{\bf Source Only} }&\includegraphics[width=\wfig\linewidth]{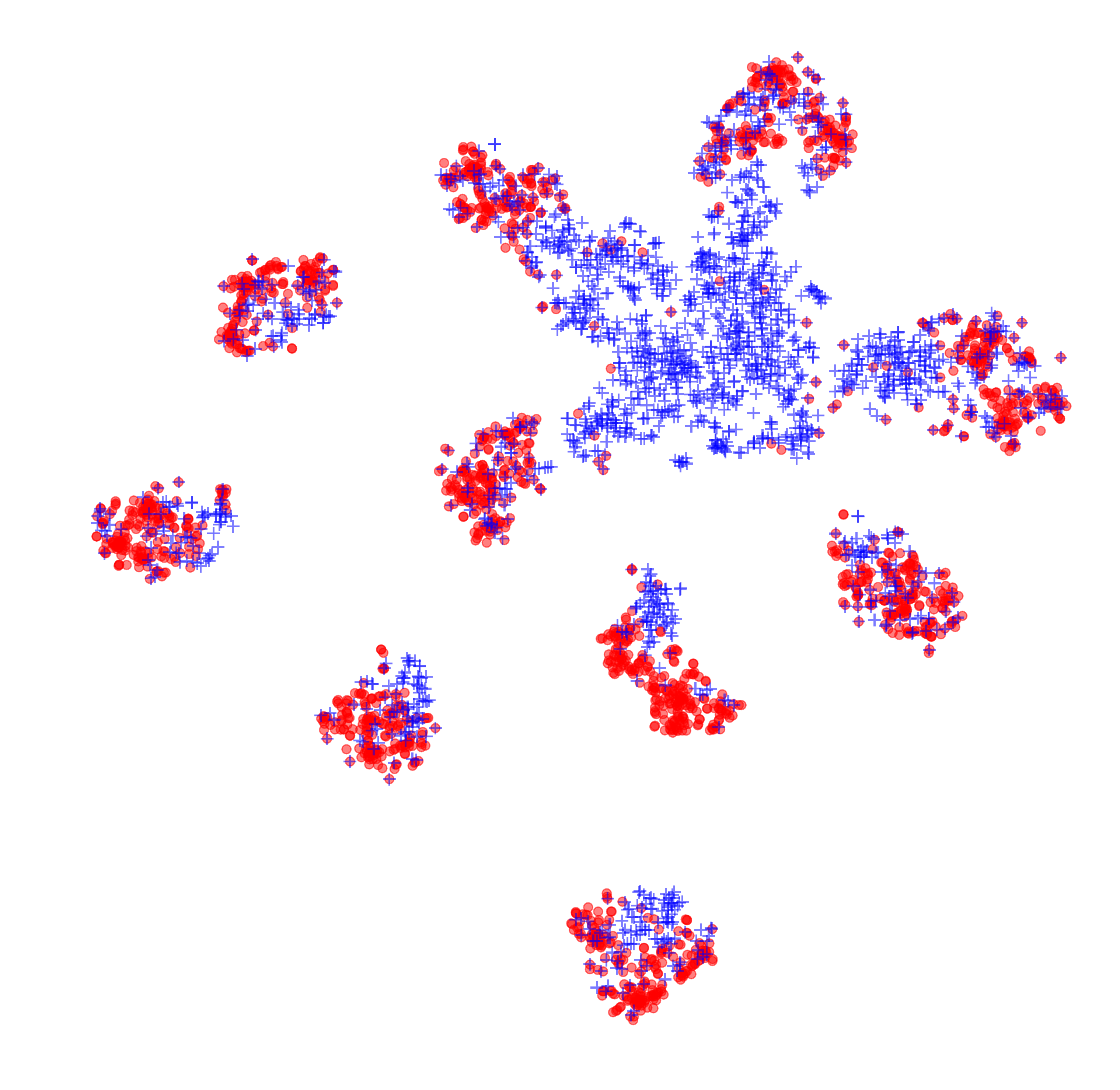} &
    \includegraphics[width=\wfig\linewidth]{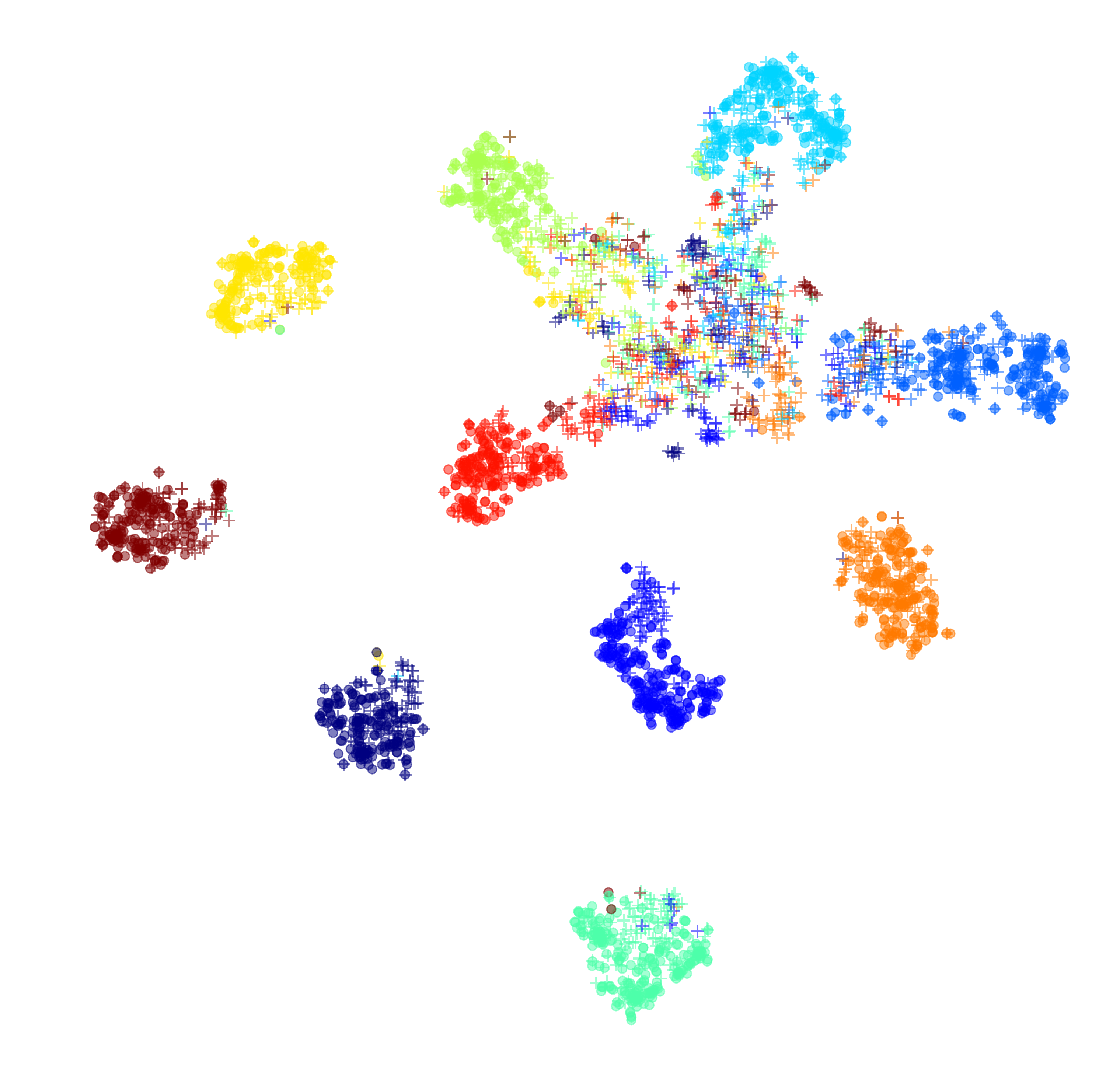}\\
    \raisebox{2.cm}{\rotatebox[origin=c]{90}{\bf DANN} }&\includegraphics[width=\wfig\linewidth]{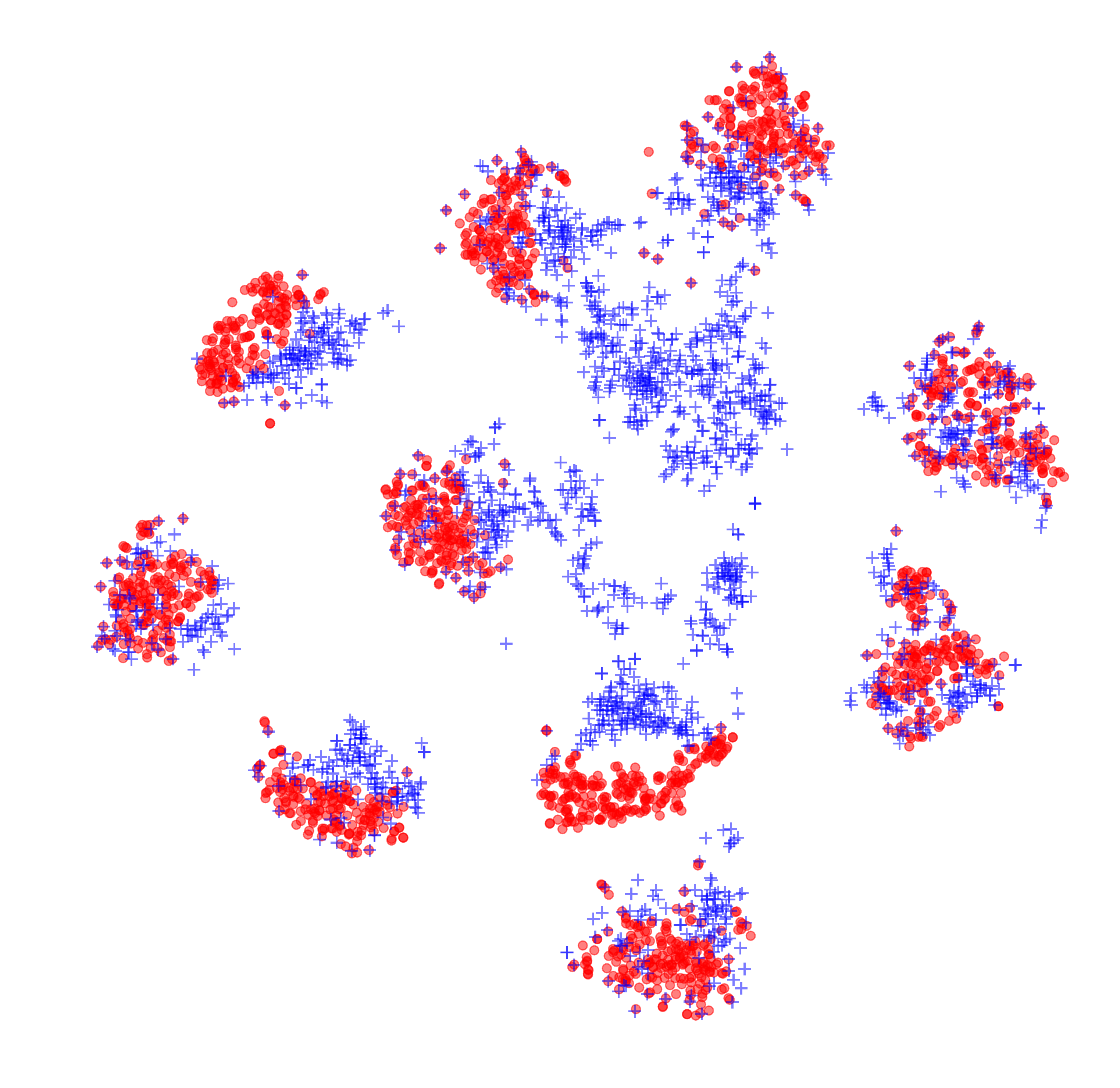} &
      \includegraphics[width=\wfig\linewidth]{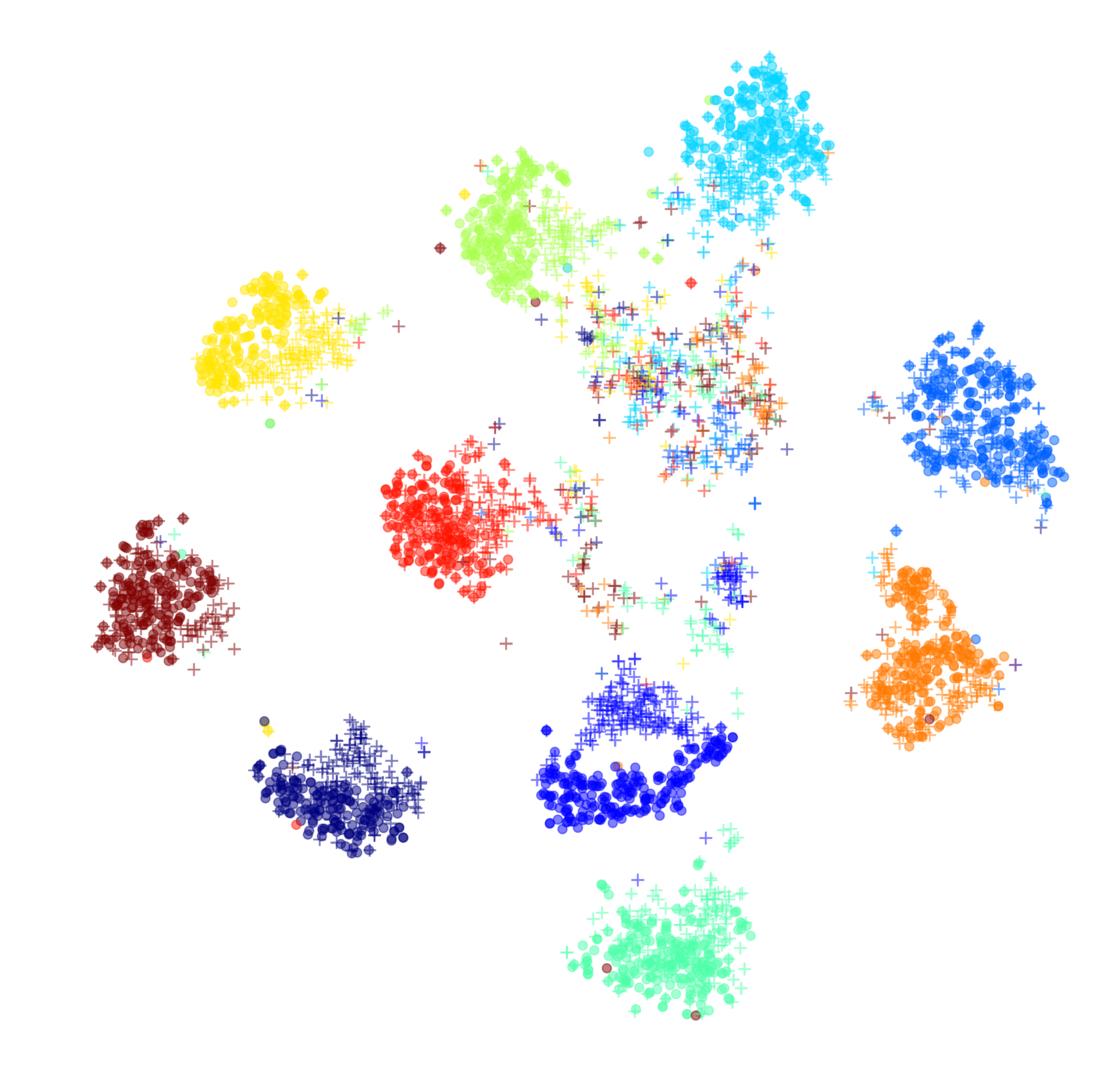}
      \\
    \raisebox{2.cm}{\rotatebox[origin=c]{90}{\bf StochJDOT} }&\includegraphics[width=\wfig\linewidth]{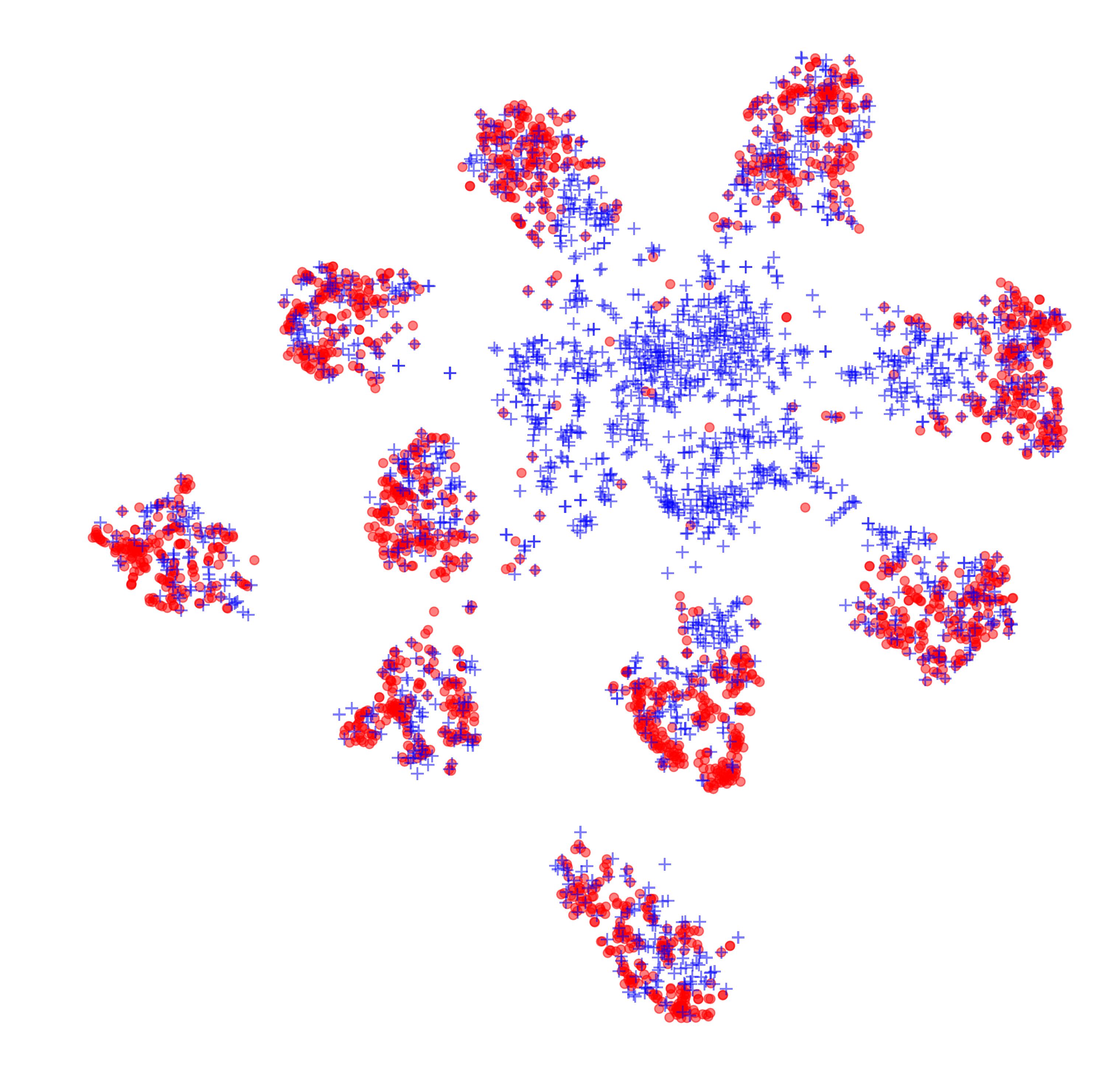} &
      \includegraphics[width=\wfig\linewidth]{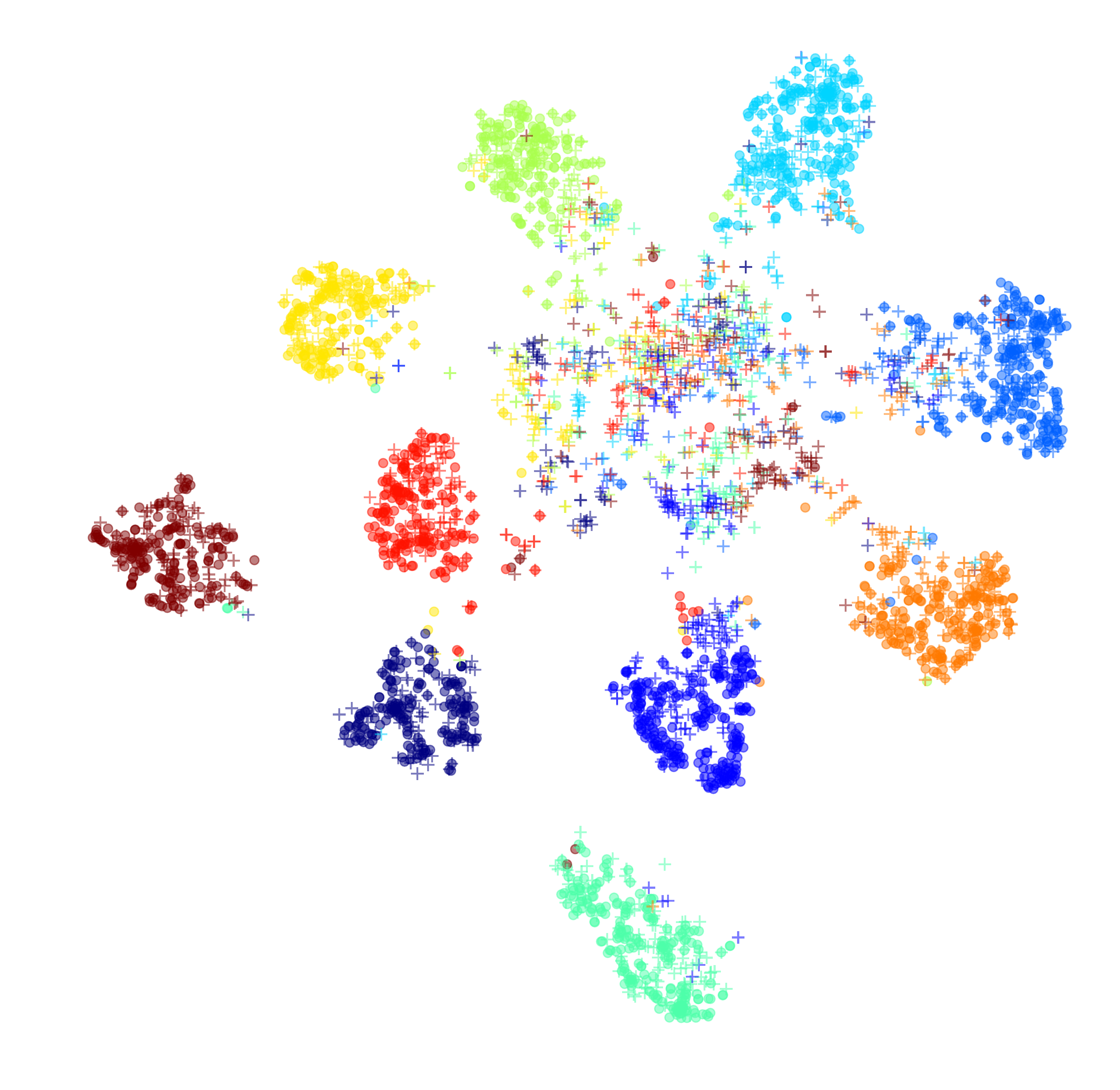}
      \\
    \raisebox{2.cm}{\rotatebox[origin=c]{90}{\bf DeepJDOT} }&\includegraphics[width=\wfig\linewidth]{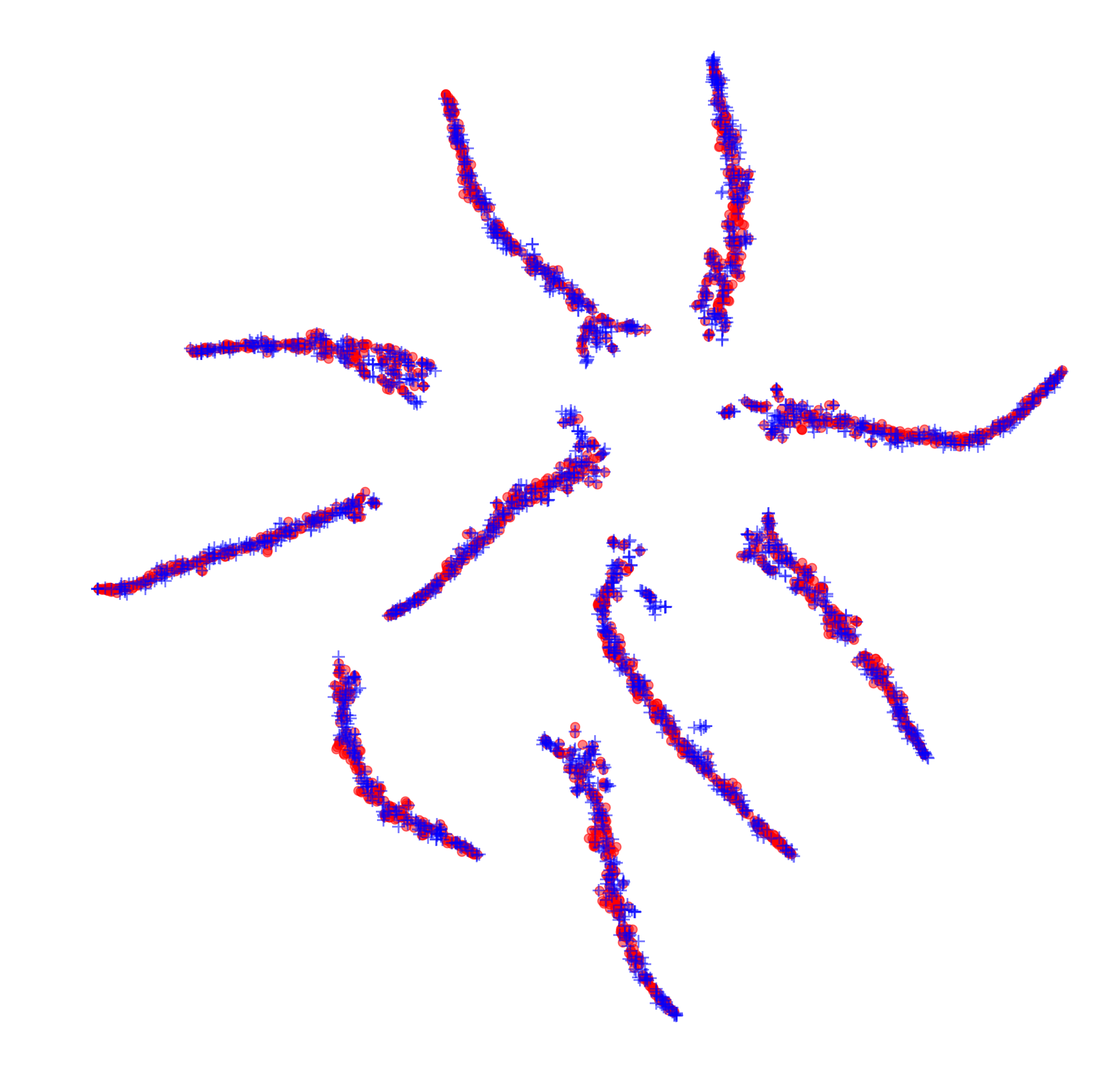} &
      \includegraphics[width=\wfig\linewidth]{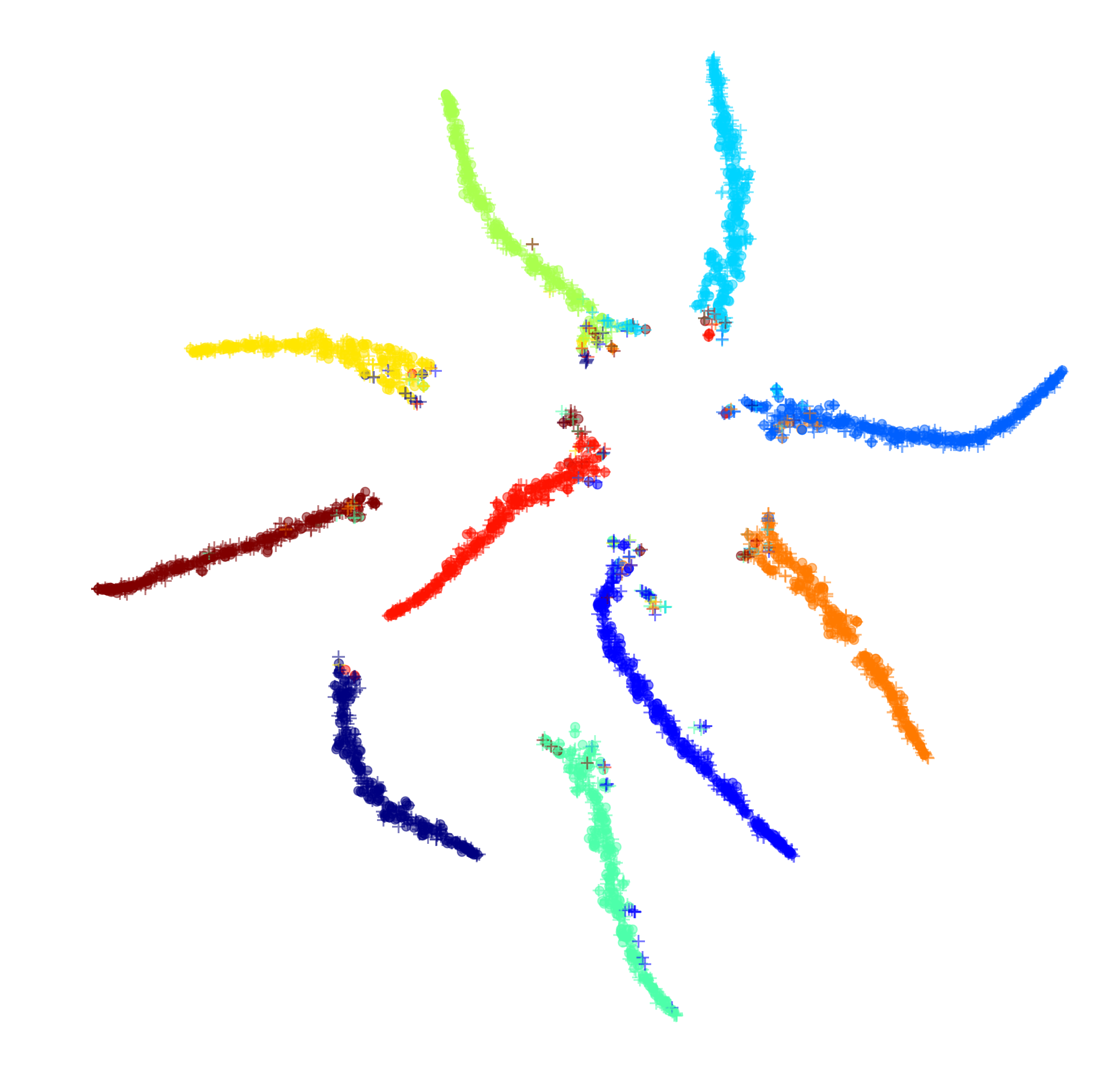}
\end{tabular}

\caption{t-SNE embeddings of 2`000 test samples for MNIST (source) and MNIST-M (target) for \texttt{Source only} classifier, \texttt{DANN}, \texttt{StochJDOT} and \texttt{DeepJDOT}. The left column shows domain comparisons, where colors represent the domain. The right column shows the ability of the methods to discriminate classes (samples are colored w.r.t. their classes).}
\label{fig:tsne}
\end{figure}

We visualize the quality of the embeddings for the source and target domain learnt by \texttt{DeepJDOT}, \texttt{StochJDOT} and \texttt{DANN} using $t$-SNE embedding on the MNIST$\rightarrow$MNIST-M adaptation task (Figure \ref{fig:tsne}). As expected, in the source model the samples from the source domain are well clustered and target samples are more scattered. The $t$-SNE embeddings with the \texttt{DANN} were not able to align the distributions well, and this observation also holds for \texttt{StochJDOT}. It is noted that \texttt{StochJDOT} does not align the distributions, but learns the classifier in target domain directly. The poor embeddings of the target samples with \texttt{StochJDOT} shows the necessity of computing the ground metric (cost function) of optimal transport in the deep CNN layers. Finally, \texttt{DeepJDOT} perfectly aligns the source domain samples and target domain samples to each other, which explains the good numerical performances reported above. The ``tentacle''-shaped and near-perfect separation of the classes in the embedding illustrate the fact that \texttt{DeepJDOT} finds an embedding that both aligns the source/target distribution, but also maximizes the margin between the classes.

\subsubsection{Ablation study} 
Table~\ref{tab:ablation} reports the results obtained in the USPS$\rightarrow$MNIST and MNIST$\rightarrow$ MNIST-M cases for models using only parts of our proposed loss (equation~\eqref{eq:jdot+source2}).
When only the JDOT loss is considered ($\alpha d+L_t$ case), the accuracy drops in both adaptation cases. This behavior might be due to overfitting of the target classifier to the noisy pseudo- (propagated) labels. However, the performance is comparable to non-adversarial discrepancy-based methods reported in Table~\ref{tab:results_largeCNN}. On the contrary, when only the feature space distribution is included in Equation~\eqref{eq:jdot+source2}, i.e. the $L_s+\alpha d$ experiment, the accuracy is close to our full model in USPS$\rightarrow$MNIST direction, but drops in the MNIST$\rightarrow$ MNIST-M one. Overall the accuracies are improved compared to the original JDOT model, which highlights the importance of including the information from the source domain. Moreover, this also highlights the importance of simultaneously updating the classifier both in the source and target domain. Summarizing, this ablation study showed that the individual components bring complimentary information to achieve the best classification results.

\begin{table}[t]
\centering
\caption{Ablation study of DeepJDOT.}
\begin{tabular}{c|c|c}
\hline
Method
  & USPS $\rightarrow$ MNIST &  MNIST $\rightarrow$ MNIST-M\\
\hline
$L_s+(\alpha d+L_t)$& 96.4  & 92.4\\
$\alpha d+L_t$  &86.41 &73.6\\
$L_s+\alpha d$  & 95.53& 82.3\\\hline
\end{tabular}
\label{tab:ablation}
\end{table}


\subsection{Office-Home}\label{sec:officehome}

\subsubsection*{Dataset} The Office-Home dataset~\cite{venkateswara2017} contains around $15'500$ images in $65$ categories from four different domains: artistic paintings, clipart, product and real-world images.

\subsubsection*{Model} In this case, we use a pre-trained VGG-16 model~\cite{simonyan2014very} with the last layer replaced, but perform no data augmentation. We use $3'250$ samples per domain to compute the optimal couplings. We compared our model with following state-of-the-art methods: CORAL\cite{Sun2016}, JDA\cite{Long2013}, DAN\cite{long15}, DANN\cite{Ganin2016}, and DAH\cite{venkateswara2017}.

\subsubsection*{Results}

Table~\ref{tab:office} lists the performance of DeepJDOT compared to a series of other adaptation methods. As can be seen, DeepJDOT outperforms all other models by a margin on all tasks, except for the adaptation from domain ``product'' to ``clipart''.

\begin{table}[t]
\centering
\caption{Performance of DeepJDOT on the Office-Home dataset. ``Ar'' = artistic paintings, ``Cl'' = clipart, ``Pr'' = product, ``Rw'' = real-world images. Performance figures of competitive methods are reported from~\cite{venkateswara2017}.} \label{tab:office}\vspace{-0.1cm}
\scriptsize{
\resizebox{\textwidth}{!}{\begin{tabular}{c|cccccccccccc|c}
\hline
Method & Ar$\rightarrow$ Cl & Ar$\rightarrow$ Pr & Ar$\rightarrow$ Rw & Cl$\rightarrow$ Ar & Cl$\rightarrow$ Pr & Cl$\rightarrow$ Rw &Pr$\rightarrow$Ar & Pr$\rightarrow$Cl&Pr$\rightarrow$Rw& Rw$\rightarrow$Ar& Rw$\rightarrow$Cl& Rw$\rightarrow$Pr& Mean  \\
\hline
CORAL\cite{Sun2016} & 27.10& 36.16 & 44.32 & 26.08 & 40.03 & 40.33& 27.77&30.54&50.61& 38.48&36.36&57.11&37.91\\
JDA \cite{Long2013} & 25.34& 35.98 & 42.94 & 24.52 & 40.19 & 40.90 & 25.96 &32.72&49.25& 35.10&35.35&55.35&36.97\\
DAN \cite{long15} & 30.66 & 42.17 & 54.13 & 32.83 &47.59 & 49.58&29.07&34.05&56.70&43.58&38.25&62.73&43.46 \\
DANN \cite{Ganin2016} & 33.33 & 42.96 & 54.42 & 32.26 & 49.13 & 49.76&30.44&38.14&56.76&44.71&42.66&64.65&44.94\\
DAH \cite{venkateswara2017} & 31.64 & 40.75 & 51.73 & 34.69 & 51.93 & 52.79&29.91&{\bf 39.63}&60.71&44.99&45.13&62.54&45.54\\
DeepJDOT & {\bf 39.73} & {\bf 50.41} & {\bf 62.49} & {\bf 39.52} & {\bf 54.35} & {\bf 53.15} & {\bf 36.72} & 39.24 & {\bf 63.55} & {\bf 52.29} &{\bf 45.43} & {\bf 70.45}& {\bf 50.67 } \\
\hline
\end{tabular}}}
\end{table}

 
\subsection{VisDA-2017}\label{sec:visda}

\subsubsection*{Dataset} The Visual Domain Adaptation classification challenge of 2017 (VisDA-2017;~\cite{visda2017}) requires training a model on renderings of 3D models for each of the 12 classes and adapting to natural images sampled from MS-COCO~\cite{lin2014microsoft} (validation set) and YouTube BoundingBoxes~\cite{real2017youtube} (test set), respectively. The test set performances reported here were evaluated on the official server.

\subsubsection*{Model} Due to VisDA's strong adaptation complexity, we employ ResNet-50~\cite{he2016deep} as a base model, replacing the last layer with two MLPs that map to 512 hidden an then to the 12 classes, respectively. We train a model on the source domain and then freeze it to calculate source feature vectors, adapting an intially identical copy to the target set. We use $4'096$ samples per domain to calculate the couplings. Data augmentation follows the scheme of \cite{french2018}.

\subsubsection*{Results} DeepJDOT's performance on VisDA-2017 is reported in Table~\ref{tab:visda} along with the baselines (DeepCORAL, DAN) from the evaluation server{\footnote{\url{\texttt{https://competitions.codalab.org/competitions/17052\#results}} \label{ft:eval_url}}}. Our entry in the evaluation server is mentioned as \texttt{oatmil}. 
We can see that our method achieved better accuracy than the distribution matching methods (DeepCORAL \cite{deepcoral}, DAN \cite{long15}) with all the classes, expect \texttt{knife}. We observe a negative transfer for the class \texttt{car} for DeepJDOT, however this phenomena is also valid with the most of the current methods (see the evaluation server results).
For a fair comparison with the rest of the methods in the evaluation server, we also showed (values in bracket of Table~\ref{tab:visda}) the accuracy difference between the source model and target model. Our method is ranked sixth when the mean accuracy is considered, and third when the difference between the source model and target model is considered at the time of publication. It is noted that the performance of our method depends on the capacity of the source model: if a larger CNN is used, the performance of our method could be improved further.

\begin{table}[t]
\centering
\caption{Performance of DeepJDOT on the VisDA 2017 classification challenge. The scores in the bracket indicate the accuracy difference between the source (unadapted) model and target (adapted) model. The respective values of CORAL and DAN are reported from the evaluation server \footref{ft:eval_url}.}\label{tab:visda}
\scriptsize{
\begin{tabular}{c|cccccccccccc|c}
\hline
Method&plane&	bcycl&	bus&	car&	horse&	knife&	mcycl&	person&	plant&	sktbd&	train&	truck & Mean \\\hline
Source only & 36.0 & 4.0& 19.9 & {\bf 94.7} & 14.8 & 0.42 & 38.7 & 3.8 & 37.4 & 8.1& 71.9 & 6.7 & 28.0 \\
DeepCORAL \cite{deepcoral} &	62.5&21.7	& 66.3&64.6	&31.1&36.7	&54.2	&24.9	&73.8	&29.9&43.4	&34.2	&45.3 (19.0) \\
DAN \cite{long15} &55.3	&18.4	&59.8&68.6	&55.3	&{\bf 41.4}	&63.4	&30.4	&78.8	&23.0	&62.9	&40.2	&49.8 (19.5) \\
DeepJDOT & {\bf 85.4} &  {\bf 50.4} & {\bf 77.3}& 87.3 & {\bf 69.1} & 14.1 & {\bf 91.5} & {\bf 53.3} & {\bf 91.9} & {\bf 31.2} & {\bf 88.5} & {\bf 61.8} &  {\bf 66.9} ({\bf 38.9}) \\
\hline
\end{tabular}}
\end{table}

\section{Conclusions}
In this paper, we proposed the DeepJDOT model for unsupervised deep domain adaptation based on optimal transport. The proposed method aims at learning a common latent space for the source and target distributions, that conveys discriminant information for both domains. This is achieved by minimizing the discrepancy of joint deep feature/labels domain distributions by means of optimal transport. We propose an efficient stochastic algorithm that solves this problem, and despite being simple and easily integrable into modern deep learning frameworks, our method outperformed the state-of-the-art on cross domain digits and office-home adaptation, and provided satisfactory results on the VisDA-2017 adaptation.

Future works will consider the evaluation of this method in multi-domains scenario, as well as more complicated cost functions taking into account similarities of the representations across the embedding layers and/or similarities of labels across different classifiers.

\section*{Acknowledgement}
This work benefited from the support of Region Bretagne grant and  OATMIL ANR-17-CE23-0012 project of the French National Research Agency (ANR). We gratefully acknowledge the support of NVIDIA Corporation with the donation of the Titan Xp GPU used for this research. The constructive comments and suggestions of anonymous reviewers are gratefully acknowledged.

\bibliographystyle{splncs}
\bibliography{egbib}
\end{document}